\documentclass[]{ceurart}

\usepackage[T1]{fontenc}
\usepackage{graphicx}
\usepackage{listings}
\usepackage{setspace}
\usepackage{xcolor}
\usepackage{amsmath}
\usepackage{hyperref}
\usepackage{xurl}
\usepackage{booktabs}
\usepackage{siunitx}
\usepackage{tikz}
\usepackage{xcolor}

\usetikzlibrary{positioning,arrows.meta,fit,calc}

\definecolor{cBox}{RGB}{225,237,250}
\definecolor{cBoxB}{RGB}{ 60,110,170}
\definecolor{cAcc}{RGB}{252,232,205}
\definecolor{cAccB}{RGB}{210,130, 40}
\definecolor{cTerm}{RGB}{210, 70, 70}
\definecolor{cSteiner}{RGB}{ 30,140, 80}
\definecolor{cArr}{RGB}{ 70, 70, 70}
\definecolor{cMute}{RGB}{170,170,170}

\definecolor{cBox}{RGB}{228,238,250}
\definecolor{cBoxB}{RGB}{ 80,130,180}
\definecolor{cCore}{RGB}{205,222,242}
\definecolor{cExt}{RGB}{245,245,245}
\definecolor{cExtB}{RGB}{160,160,160}
\definecolor{cAcc}{RGB}{252,232,205}
\definecolor{cAccB}{RGB}{210,150, 80}
\definecolor{cArr}{RGB}{ 80, 80, 80}
\definecolor{cOff}{RGB}{160,160,160}

\def\stagew{78mm}
\def\stageh{22mm}
\def\divx{42mm}

\begin{document}
\title{Making Mathematical Knowledge Explainable, Accessible and Interoperable Through Large Language Model Integration}

%
%




\copyrightyear{2026}
\copyrightclause{Copyright for this paper by its authors. Use permitted under Creative Commons License Attribution 4.0 International (CC BY 4.0).}
\conference{Wikidata'26: Wikidata Workshop at ISWC 2026}

\author[1]{Jan Range}[
orcid=0000-0001-6478-1051,
email=jan.range@simtech.uni-stuttgart.de,
]

\address[1]{Stuttgart Center for Simulation Science, University of Stuttgart, Germany}

\author[2]{Bj\"{o}rn Schembera}[
orcid=0000-0003-2860-6621,
email=bjoern.schembera@mathematik.uni-stuttgart.de,
]
\cormark[1]
\address[2]{Institute of Applied Analysis and Numerical Simulation, University of Stuttgart, Allmandring 5b, 70569 Stuttgart, Germany}

\author[1,2]{Dominik G\"{o}ddeke}[
orcid=0000-0002-1552-497X,
email=dominik.goeddeke@mathematik.uni-stuttgart.de,
]

\cortext[1]{Corresponding author.}

\begin{abstract}
Mathematical models are central to formalizing research problems, yet their documentation often falls short of FAIR principles. Knowledge bases such as the Mathematical Model Database (MathModDB) address this gap by providing curated, semantically rich representations of mathematical models. Built on Wikibase, the same open-source infrastructure underlying Wikidata, MathModDB utilizes Semantic Web technologies to support Linked Open Data, collaborative editing, and the storage of semantically enriched metadata, making it a domain-specific knowledge graph within the broader Wikidata ecosystem.
However, access to MathModDB currently requires either navigating a complex web interface or proficiency in SPARQL and Wikibase APIs, posing significant barriers for potential users. In addition, the combination of such curated knowledge bases with actual research data stored, e.g., in Dataverse repository instances, remains a challenge.
To overcome these limitations, we propose integrating Large Language Models (LLMs) with MathModDB via a Model Context Protocol (MCP) server that exposes a vector-indexed schema retrieval and Steiner-tree-based join planner, combining dialogue-based natural language interaction with curated, epistemically grounded knowledge. Although instantiated on MathModDB, the architecture can be applied to other Wikibase-based systems. We demonstrate that this approach enables epistemically grounded LLM usage, improves model explainability and accessibility beyond what the standard Wikibase interface offers, and simplifies interoperability with external databases and tools, such as Dataverse data repositories. We illustrate the benefits of combining the accessibility of an LLM with the epistemic safety of a curated knowledge base through the adaptability of the MCP protocol by two use cases: linking an enzyme-kinetics dataset from a Dataverse repository to its mathematical model in MathModDB, and exploring the Stokes-Darcy model for coupled flow including federated retrieval of creator biographies from Wikidata. 
\end{abstract}


 \maketitle   

\section{Introduction}

Mathematical knowledge is vast, ranging from classic mathematical artifacts like proofs on paper to complex numerical models on supercomputers, and it is usually scattered through different categories of data, i.e., symbolic, numeric, geometric, observational, model, or text data~\cite{Conrad2024,Hulek2019,Schubotz2023}. For the majority of scientific domains, mathematical models are the central means to describe their research problems in a formalized and abstract way, so researchers can make predictions, claims, and analyzes about the research problem.

To comply with the FAIR principles~\cite{Wilkinson2016} and good scientific practice, all steps that led to a research result (i.e., the predictions, claims and analyses) must be documented. This includes, in particular, documentation of the mathematical models used, created or modified during the research process. Using mathematical knowledge bases for documenting the involved  mathematical models enables exploring, researching and finding similar, alternate or complementing models and solution schemes. Mathematical knowledge bases for mathematical models exist, and one example is the Mathematical Model Database (MathModDB)~\cite{Schembera2025}. It is a curated database of mathematical models, developed by the Mathematical Research Data Initiative (MaRDI)~\cite{MaRDI2022}. MaRDI is embedded into the greater national research data infrastructure (NFDI) in Germany, giving MathModDB a potentially high impact in the German research landscape and beyond. MathModDB uses Wikibase~\cite{Vrandevcic2014} as its technology stack -- the same open-source infrastructure underlying Wikidata -- fostering a flexible, semantic, and extensible representation of data assets in line with Semantic Web standards~\cite{BernersLee2001}. Its support for Linked Open Data, collaborative editing, and scalability makes it a foundation for managing complex mathematical models, and positions MathModDB as a domain-specific extension of the broader Wikidata ecosystem.
MathModDB with Wikibase as its technological platform is in principle explainable AI ready (XAIR) as it consists of data that are readable by both humans and machines, and which enables the storage of epistemic metadata~\cite{Horsch2023} about mathematical models (e.g., descriptions, governing equations, assumptions, constraints, conditions, quantities, but also links to proofs or other theoretic grounding), going beyond mere provenance metadata. However, usage of MathModDB is either limited to a web interface for browsing and modifying the data, or requires expertise in writing API calls or SPARQL for complex, federated queries. This is often particularly discouraging for users, as it means they have to learn a new and formal query language or the technical details of the Wikibase API. For machine interaction, too, programmers must be proficient in SPARQL. Results returned are in machine-readable formats such as JSON, XML or CSV, which require further processing to make them explainable, presentable and thus fully XAIR~\cite{Schembera2025_DIKW}.

To address these shortcomings and as users prefer natural-language queries over writing SPARQL, we propose a Model Context Protocol (MCP) based approach to combine the strengths of LLMs with curated knowledge bases. Although instantiated on MathModDB, the architecture is generic across Wikibase-based systems (including Wikidata) and thus offers a reusable integration pattern for the broader Wikibase and Linked Open Data ecosystem.

In this paper, we demonstrate that curated knowledge stored in the MathModDB knowledge graph supports epistemically safe use of LLMs and facilitates steps toward XAIR-compliant research outputs in a Semantic Web setting. We show that a dialogue-based, LLM-supported interface offers substantially richer access to curated mathematical knowledge than the standard Wikibase interface -- in particular with respect to model explainability and navigability.
We quantify our approach by schema-retrieval evaluation and illustrate its application through two use cases. In the first use case, we show that the MCP-based approach enables end-to-end interoperability between research data and mathematical knowledge: using only a natural language query, an agent links an enzyme kinetics dataset deposited in a Dataverse data repository to its corresponding mathematical model in MathModDB, identifying the underlying rate laws and mechanistic model class without any manual integration effort. In the second case, we use the Stokes Darcy model for coupled flow to demonstrate that even structurally complex models can be explored reliably: the correct coupling conditions are retrieved directly, and biographical information about the creators, spread across different knowledge bases, is retrieved without requiring the user to construct federated SPARQL queries or possess detailed knowledge of the underlying LOD database structures.

\section{Related Work}

\subsection{Knowledge Bases for Mathematical Models}
\label{subsec:KnowledgeBases}

Mathematical knowledge management was already being discussed in the early 2000s~\cite{Farmer2004}. Knowledge bases for mathematical models include databases, ontologies and taxonomic classifications and are discussed in this subsection. 

Ontologies for mathematical models are either domain-specific~\cite{Chelliah2013,Inizan2021,Suresh2008,Suresh2010,Snytnikov2020} or primarily taxonomic in nature~\cite{Falileeva2020,Kirillovich2021}, and none provides a unified, general and semantically queryable representation of mathematical relationships enriched with epistemic metadata. 
MathModDB addresses this gap. As part of the NFDI project MaRDI, MathModDB was initially developed as an ontology~\cite{Schembera2024} for mathematical models and subsequently populated as a knowledge graph~\cite{Schembera2025}. MathModDB consists of 8 classes for a rich semantic description of mathematical models in their specific research context: \textit{Academic Discipline}, \textit{Research Problem}, \textit{Mathematical Model}, \textit{Mathematical Formulation}, \textit{Quantity}, \textit{Quantity Kind}, \textit{Computational Task} and \textit{Publication}. By February 2025, MathModDB was moved from the OWL world to the Wikibase ecosystem~\cite{schembera_2025_CoRDI}, becoming a subgraph of the larger MaRDI KG~\cite{Schubotz2023}. The Wikibase technology stack offers open-source development momentum, a large mathematical user base~\cite{Scharpf2021}, and the possibility to build on existing knowledge from Wikidata and other external knowledge bases. This potential is already being realized: relevant entities from Wikidata, such as publications, persons, and research institutions, have been imported into MathModDB, and direct links to Wikidata have been set for many data items in MathModDB. Mathematics-specific extensions include rendering of mathematical expressions, formula search, and the integration of data sources such as zbMath and swMath. Moreover, Wikibase supports qualifiers for relations -- important in mathematics to denote special equations such as assumptions or conditions -- and facilitates linking with the broader MaRDI KG and external knowledge graphs following LOD principles. As of July 2026, MathModDB includes 229 curated mathematical models and 2,153 entries comprising 26,041 statements in total.

\subsection{MCP for Knowledge Graphs and Research Data}

The Model Context Protocol (MCP), introduced by Anthropic in late 2024, serves as an open standard for bidirectional and typed communication between large language model (LLM) agents and external tools or data sources. MCP has rapidly emerged as the primary integration layer for agentic AI~\cite{Hou2025,Ray2025}. In the context of the Semantic Web, MCP is increasingly regarded as a suitable framework for knowledge graphs and SPARQL endpoints, given their standardized query languages, metadata, and inherent support for federation~\cite{Dobriy2026}.

Multiple MCP servers for knowledge graphs are currently available. 
The Wikidata community maintains a Wikidata MCP server\footnote{\url{https://www.wikidata.org/wiki/Wikidata:MCP}} that provides both SPARQL execution and a vector index over Wikidata items and properties to enable semantic search. The MathModDB MCP can be understood as a domain-specific complement: where the Wikidata MCP server offers broad general-purpose knowledge, the MathModDB MCP server provides epistemically curated mathematical knowledge, grounded in formal model descriptions and governed by domain experts.
Dobriy et al.~\cite{Dobriy2026} extend the Spider4SPARQL benchmark to federated knowledge graph question answering (KGQA) and evaluate SPARQL-MCP agents in the areas of endpoint discovery, VoID-based schema exploration, and query formulation. Their findings indicate that high-level natural-language endpoint descriptions are more effective than structured VoID metadata for guiding source selection. The optimal structuring of schema metadata for agents remains an open research question. Schema-aware NL-to-SPARQL translation has prior history in KGQA. Bio-SODA~\cite{BioSODA2021} constructs SPARQL candidates by solving an approximate Steiner tree problem over a schema graph -- the closest structural predecessor to our join planner -- but without LLM integration or MCP. Emonet et al.~\cite{Emonet2025} use LLM-based RAG with schema retrieval over federated endpoints, but rely on post-hoc query correction rather than a generation-time prior. Recently, SAGA~\cite{SAGA2026} proposed schema-aware grounding for agentic SPARQL generation on Wikidata. In contrast, our approach combines Steiner-tree join planning with dual-encoder retrieval within a stateless MCP interface, targeting a Wikibase-based expert-curated knowledge graph with qualifier-mediated triple patterns.
In related domains, MCP has been adapted for research data infrastructure: Pan et al.~\cite{Pan2025} unify heterogeneous research cyberinfrastructure, Dasgupta et al.~\cite{Dasgupta2025} expose a DSpace digital repository, and Kuehl et al.~\cite{Kuehl2025} establish a community registry of MCP servers for biomedical knowledge bases.

\subsection{FAIR, XAIR, and epistemic Metadata}
\label{subsec:XAIR}

For around a decade, the FAIR principles have shaped best practices in research data management~\cite{Wilkinson2016}. They address data provenance and semantics, ensuring that the origin and meaning of data are documented in a machine-readable way. However, for applications that require scientific rigor and reproducibility, knowing what data means is not sufficient -- one must also be able to justify why it is trustworthy. The transition from FAIR to XAIR demands metadata annotations that go beyond semantic description and document the epistemic status of data and derived claims~\cite{Horsch2023european}: the assumptions underlying a model, the conditions under which it is valid, or the reasoning that connects a mathematical formulation to a physical phenomenon. This shift from a semantic to an epistemic perspective is operationalized through epistemic metadata~\cite{Horsch2023}.
MathModDB is designed around precisely this concept: it stores epistemic metadata about mathematical models -- including their governing equations, coupling conditions, and formal relationships -- in a curated, machine-actionable knowledge graph, making it a XAIR-compliant resource for mathematical knowledge.

\section{System Architecture}\label{sec:architecture}

As described in Sec.~\ref{subsec:KnowledgeBases}, MathModDB is a curated, Wikibase-based knowledge base of mathematical models. It is human- and machine-readable, carries rich epistemic metadata, and is therefore XAIR-compliant (c.f. Sec.~\ref{subsec:XAIR}). However, its practical utility is limited by access barriers: Users must navigate the web interface manually, write SPARQL queries, or engage Wikibase APIs directly, which requires specialist knowledge. This is a recognized shortcoming~\cite{Schubotz2023} and a common challenge across Wikibase-based systems, especially for federated queries spanning multiple LOD databases~\cite{Liu2024}. 
To overcome these barriers and extend MathModDB with dialogue-based interaction and the contextual understanding of an LLM, we introduce the MathModDB MCP server. The server architecture is not specific to MathModDB: any Wikibase instance that exposes a SPARQL endpoint and an indexable schema can be wrapped with the same approach, making this a reusable integration pattern for Wikibase-based knowledge graphs. The transfer requires only a SPARQL endpoint exposing the target graph and an indexable schema; no modifications to the underlying knowledge graph are needed.

\begin{figure}[!h]
  \centering
  \begin{tikzpicture}[
  font=\sffamily\small,
  >={Stealth[length=2mm]},
  ext/.style    ={draw=cExtB,fill=cExt,rounded corners=1.5pt,
                  align=center,inner sep=3pt,line width=0.4pt,
                  minimum width=32mm,minimum height=10mm},
  tool/.style   ={draw=cBoxB,fill=cBox,rounded corners=1.5pt,
                  align=center,inner sep=3pt,line width=0.4pt,
                  minimum width=32mm,minimum height=8mm},
  core/.style   ={draw=cBoxB,fill=cCore,rounded corners=1.5pt,
                  align=center,inner sep=3pt,line width=0.6pt,
                  minimum width=32mm,minimum height=10mm},
  store/.style  ={draw=cAccB,fill=cAcc,rounded corners=1.5pt,
                  align=center,inner sep=2.5pt,line width=0.4pt,
                  minimum height=8mm},
  off/.style    ={draw=cOff,rounded corners=1.5pt,
                  align=center,inner sep=2.5pt,line width=0.3pt,
                  dashed,text=cOff!50!black,
                  font=\sffamily\footnotesize,minimum height=8mm},
  arr/.style    ={->,line width=0.5pt,draw=cArr,
                  rounded corners=1.5pt},
  bidir/.style  ={<->,line width=0.5pt,draw=cArr,
                  rounded corners=1.5pt},
  offarr/.style ={->,line width=0.5pt,dashed,draw=cOff,
                  rounded corners=1.5pt},
  lbl/.style    ={font=\sffamily\footnotesize\itshape,inner sep=1pt,text=cArr}
]
 
\node[ext,minimum width=28mm,minimum height=8mm] (agent) {LLM agent};
 
\node[tool,below=10mm of agent,xshift=-19mm] (sparql)
  {\texttt{(Batch\_)SPARQL\_Query}};
\node[tool,below=10mm of agent,xshift= 19mm] (explore)
  {\texttt{Explore\_Ontology}};
 
\node[ext,below=12mm of sparql] (kb)
  {MathModDB SPARQL\\[-1pt]\footnotesize endpoint};
\node[core,below=12mm of explore] (planner)
  {Planner\\[-1pt]\footnotesize schema graph $+$ Steiner};
\node[store,right=4mm of planner] (qdrant) {Qdrant};
\node[off,right=5mm of qdrant] (idx)
  {Schema indexer\\[-1pt](offline)};
 
\coordinate (mcpA) at ([xshift=-2mm, yshift= 2mm] sparql.north west);
\coordinate (mcpB) at ([xshift= 2mm, yshift= 2mm] explore.north east);
\coordinate (mcpC) at ([xshift= 2mm, yshift=-2mm] planner.south east);
\coordinate (mcpD) at ([xshift=-2mm, yshift=-2mm] planner.south west);
\coordinate (mcpE) at ([xshift=-2mm, yshift=-2mm] planner.south west |- sparql.south);
\coordinate (mcpF) at ([xshift=-2mm, yshift=-2mm] sparql.south west);
 
\draw[draw=cBoxB!55,dashed,rounded corners=2pt]
  (mcpA) -- (mcpB) -- (mcpC) -- (mcpD) -- (mcpE) -- (mcpF) -- cycle;
 
\node[font=\sffamily\footnotesize\bfseries,text=cBoxB,inner sep=1pt,anchor=south]
  at ($(mcpA)!0.5!(mcpB) + (0,0.5mm)$) {FastMCP server};
 
\draw[bidir] (agent.south) -- ++(0,-3mm) -| (sparql.north);
\draw[bidir] (agent.south) -- ++(0,-3mm) -| (explore.north);
 
\draw[arr]   (sparql.south)  -- (kb.north);
\draw[bidir] (explore.south) -- node[lbl,right=0pt]{TOON subschema} (planner.north);
 
\draw[bidir] (planner.east) -- (qdrant.west);
 
\draw[offarr] (idx.west) -- (qdrant.east);
\draw[offarr] (idx.south) |- ($(kb.south)+(0,-5mm)$) -| (kb.south);
 
\end{tikzpicture}
  \caption{Architecture of the MathModDB MCP server. The LLM agent
interacts with three tools exposed by the FastMCP server (dashed
outline): \texttt{Explore\_Ontology} delegates to the Planner, which
retrieves and re-ranks schema candidates from the offline-built Qdrant
index and returns a TOON-encoded subschema. \texttt{SPARQL\_Query} and
\texttt{Batch\_SPARQL\_Query} are direct passthroughs to the MathModDB
endpoint. Solid arrows denote runtime flow and dashed arrows denote the
offline path, in which the Schema indexer harvests the endpoint and
populates Qdrant.}
  \label{fig:stack}
\end{figure}
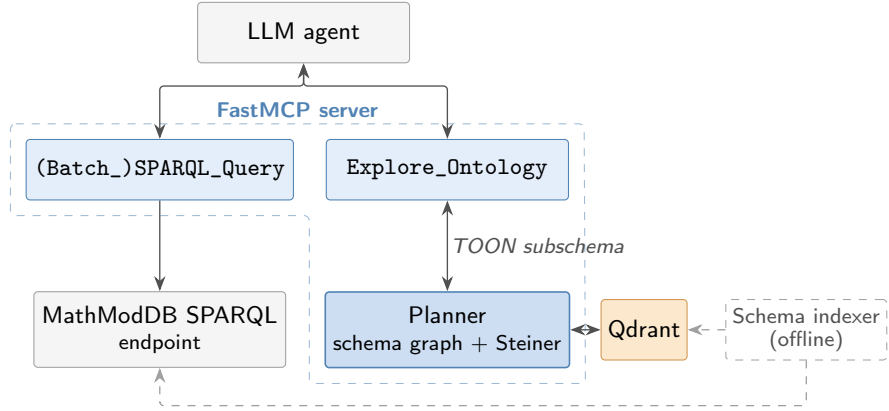

The server comprises three components: an offline schema index built once from the MathModDB Wikibase instance, a runtime query-planning pipeline executed per request, and an MCP interface exposing this pipeline to LLM agents via network transport. In this setup, MathModDB serves as contextual grounding for LLM-based systems~\cite{Abu2024,Sequeda2025}, providing the semantic and epistemic trust required in scientific settings. The system architecture is illustrated in Fig.~\ref{fig:stack}. The source code of the MathMod MCP server is referenced in Sec.~\ref{sec:SupplementaryMaterial}, Supplementary Material.

\subsection{Schema Indexer}

During initialization, the server queries the MathModDB SPARQL endpoint at \url{https://query.portal.mardi4nfdi.de/sparql} to collect all entities marked with the MathModDB corpus identifier (\texttt{Q6534265}). For each class, the server retrieves its Wikibase identifier, English label and description, associated object and data properties, and observed usage counts. For each object property, empirical subject-to-object class connections and their frequencies are recorded through separate SPARQL queries. Qualifier properties are extracted along with their usage paths, represented as subject-class and main-property pairs. The resulting structure is serialized to a YAML cache and reused across server restarts. This process produces four typed entity sets: \textit{classes}, \textit{object properties}, \textit{data properties}, and \textit{qualifier properties}. Each set contains a Wikibase \texttt{Q}/\texttt{P} identifier, textual metadata, and relational metadata for downstream graph construction.

For semantic retrieval, each entity is represented as a retrieval string derived from its label and description, optionally enriched with an LLM-generated summary. These strings are embedded using two complementary encoders: A dense OpenAI \textit{text-embedding-3-large}\footnote{\url{https://developers.openai.com/api/docs/models/text-embedding-3-large}} model for high-recall candidate generation and a ColBERT v2 \cite{ColBert} late-interaction model for fine-grained re-ranking. The resulting vectors are stored in four typed Qdrant collections, one for each entity type, on disk alongside the complete entity payload. These collections persist across server restarts and are rebuilt only when an explicit refresh flag is set. For the reported experiments, entity descriptions were enriched with LLM-generated summaries, which augments each entity's retrieval string with a GPT-generated paragraph optimized for vector search.

\subsection{Query Executioner}

During query execution, the server executes the complete query-planning pipeline described in Section~\ref{sec:methods} as a single synchronous operation. The pipeline processes a natural-language input, performs dual-encoder retrieval across all four Qdrant collections, constructs and conditions the schema graph, computes the Steiner subgraph, and attaches qualifier and data-property constraints to generate a TOON-encoded output payload. All intermediate states are maintained in memory for the duration of the request, with no per-request persistence.

\subsection{MCP Server Interface}

The MCP server component is implemented in Python using FastMCP and is served over streamable HTTP, ensuring compatibility with any MCP-capable LLM client. Three tools are registered: \texttt{Explore\_Ontology}, which accepts a short natural-language query and returns a TOON-encoded payload containing ranked schema candidates and the Steiner-derived join blueprint, and must be invoked prior to any SPARQL execution. \texttt{SPARQL\_Query}, which executes a single SPARQL \texttt{SELECT} query against the MathModDB endpoint with preconfigured \texttt{wd:} and \texttt{wdt:} prefix bindings, subject to a strict result limit of 100 per query. Finally, \texttt{Batch\_SPARQL\_Query}, which executes a named dictionary of SPARQL queries in a single call, returning results keyed by the caller-supplied names. The recommended agent workflow is schema-first: the LLM first invokes \texttt{Explore\_Ontology} to obtain valid entity identifiers and a join blueprint before issuing targeted data retrieval queries. The TOON-encoded subschema returned by \texttt{Explore\_Ontology} acts as a generation-time prior, fixing the candidate class and predicate space before any query is written rather than filtering results afterwards.

\section{Methods}

\label{sec:methods}

\begin{figure}[!h]
  \centering
  \begin{tikzpicture}[
  font=\sffamily\small,
  >={Stealth[length=2.4mm]},
  stage/.style ={draw=cBoxB,fill=cBox,rounded corners=2pt,
                 line width=0.5pt,
                 minimum width=\stagew,minimum height=\stageh,
                 inner sep=0pt},
  io/.style    ={draw=cAccB,fill=cAcc,rounded corners=2pt,
                 align=center,inner sep=4pt,line width=0.5pt,
                 minimum width=34mm,minimum height=8mm},
  big/.style   ={->,draw=cArr,line width=0.7pt},
  hd/.style    ={font=\sffamily\small\bfseries,text=cBoxB},
  bd/.style    ={font=\sffamily\footnotesize}
]

\node[io] (q) {NL query $q$};

\node[stage,below=4mm of q] (s1) {};
\node[hd,anchor=north west] at ($(s1.north west)+(2.5mm,-2mm)$)
  {1. Candidate retrieval};
\node[bd,anchor=north west,align=left,text width=38mm]
  at ($(s1.north west)+(2.5mm,-7mm)$)
  {dense $f_d$ + ColBERT\\
   per typed Qdrant collection\\
   $\to$ top-$k$, $\hat{s}\!\in\![0,1]$};
\coordinate (s1c) at ($(s1.east)+(-18mm,0)$);
\node[bd,text=cArr] at ($(s1c)+(0,7mm)$) {$q$};
\foreach \i/\lbltxt in {1/cls, 2/obj, 3/dat, 4/qua} {
  \node[draw=cAccB!70,fill=cAcc!60,rounded corners=1pt,
        minimum width=6mm,minimum height=3.5mm,inner sep=0.5pt,
        font=\sffamily\scriptsize]
    (col\i) at ($(s1c)+({(\i-2.5)*7.5mm},-1mm)$) {\strut\lbltxt};
  \draw[->,draw=cArr!70,line width=0.4pt]
    ($(s1c)+(0,5mm)$) -- (col\i.north);
}

\node[stage,below=4mm of s1] (s2) {};
\node[hd,anchor=north west] at ($(s2.north west)+(2.5mm,-2mm)$)
  {2. Edge weighting};
\node[bd,anchor=north west,align=left,text width=38mm]
  at ($(s2.north west)+(2.5mm,-7mm)$)
  {$w(e)\!=\!\omega_r f_\mathrm{sim} f_\mathrm{freq} f_\mathrm{hub}$\\[1pt]
   relevance, frequency, hub\\
   penalties $\to$ cost landscape};
\coordinate (s2c) at ($(s2.east)+(-18mm,0)$);
\coordinate (g2a) at ($(s2c)+(-9mm, 5mm)$);
\coordinate (g2b) at ($(s2c)+( 9mm, 5mm)$);
\coordinate (g2c) at ($(s2c)+(-9mm,-5mm)$);
\coordinate (g2d) at ($(s2c)+( 9mm,-5mm)$);
\coordinate (g2e) at ($(s2c)+( 0  , 0  )$);
\draw[line width=1.6pt,draw=cBoxB] (g2a) -- (g2e);
\draw[line width=1.6pt,draw=cBoxB] (g2e) -- (g2d);
\draw[line width=0.5pt,draw=cMute] (g2a) -- (g2b);
\draw[line width=0.5pt,draw=cMute] (g2b) -- (g2d);
\draw[line width=0.5pt,draw=cMute] (g2c) -- (g2e);
\draw[line width=0.5pt,draw=cMute] (g2c) -- (g2d);
\foreach \n in {g2a,g2b,g2c,g2d,g2e} { \fill[cBoxB] (\n) circle (1.1mm); }

\node[stage,below=4mm of s2] (s3) {};
\node[hd,anchor=north west] at ($(s3.north west)+(2.5mm,-2mm)$)
  {3. Steiner-tree planner};
\node[bd,anchor=north west,align=left,text width=38mm]
  at ($(s3.north west)+(2.5mm,-7mm)$)
  {min-weight subtree $\mathcal{S}$\\[1pt]
   spans retrieved terminals\\
   {\color{cTerm}$\bullet$}\,terminal {\color{cBoxB}$\bullet$}\,Steiner pt.};
\coordinate (s3c) at ($(s3.east)+(-18mm,0)$);
\coordinate (g3a) at ($(s3c)+(-9mm, 5mm)$);
\coordinate (g3b) at ($(s3c)+( 9mm, 5mm)$);
\coordinate (g3c) at ($(s3c)+(-9mm,-5mm)$);
\coordinate (g3d) at ($(s3c)+( 9mm,-5mm)$);
\coordinate (g3e) at ($(s3c)+( 0  , 0  )$);
\draw[line width=0.5pt,draw=cMute!60] (g3a) -- (g3b);
\draw[line width=0.5pt,draw=cMute!60] (g3b) -- (g3d);
\draw[line width=0.5pt,draw=cMute!60] (g3c) -- (g3d);
\draw[line width=0.5pt,draw=cMute!60] (g3c) -- (g3e);
\draw[line width=1.4pt,draw=cSteiner] (g3a) -- (g3e);
\draw[line width=1.4pt,draw=cSteiner] (g3e) -- (g3d);
\fill[cTerm] (g3a) circle (1.4mm);
\fill[cTerm] (g3d) circle (1.4mm);
\fill[cBoxB] (g3e) circle (1.1mm);
\fill[cMute] (g3b) circle (1.0mm);
\fill[cMute] (g3c) circle (1.0mm);

\node[stage,below=4mm of s3] (s4) {};
\node[hd,anchor=north west] at ($(s4.north west)+(2.5mm,-2mm)$)
  {4. Linearize \& attach};
\node[bd,anchor=north west,align=left,text width=38mm]
  at ($(s4.north west)+(2.5mm,-7mm)$)
  {triple patterns $(c_i, r, c_j)$\\[1pt]
   {\color{cAccB}$+$ data props (SELECT)}\\
   {\color{cSteiner}$+$ qualifier patterns}};
\coordinate (s4c) at ($(s4.east)+(-18mm,0)$);
\node[draw=cBoxB,fill=white,rounded corners=1pt,inner sep=2pt,
      minimum width=22mm,minimum height=5mm,
      font=\sffamily\scriptsize]
  at ($(s4c)+(0,5mm)$) {$(c_i,\,r,\,c_j)$};
\node[draw=cAccB,fill=cAcc!60,rounded corners=1pt,inner sep=2pt,
      minimum width=22mm,minimum height=4mm,
      font=\sffamily\scriptsize,text=cAccB]
  at ($(s4c)+(0,0mm)$) {data props};
\node[draw=cSteiner,fill=cSteiner!15,rounded corners=1pt,inner sep=2pt,
      minimum width=22mm,minimum height=4mm,
      font=\sffamily\scriptsize,text=cSteiner]
  at ($(s4c)+(0,-5mm)$) {qualifiers};

\node[io,below=4mm of s4] (out) {TOON $\to$ agent};

\draw[big] (q)  -- (s1);
\draw[big] (s1) -- (s2);
\draw[big] (s2) -- (s3);
\draw[big] (s3) -- (s4);
\draw[big] (s4) -- (out);

\foreach \s in {s1,s2,s3,s4} {
  \draw[draw=cBoxB!30,line width=0.3pt]
    ($(\s.north west)+(\divx,-2mm)$) -- ($(\s.south west)+(\divx,2mm)$);
}

\end{tikzpicture}
  \caption{Query-planning pipeline executed by \texttt{Explore\_Ontology}.
  A natural-language query~$q$ is retrieved against the four typed Qdrant
  collections via a dense encoder $f_d$ followed by ColBERT re-ranking
  (Section~\ref{sec:retrieval}). The normalized scores condition the
  schema graph through relevance, frequency, and hub penalties.
  Steiner tree (Section~\ref{sec:steiner}) selects the minimum-weight
  subtree $\mathcal{S}$ spanning all retrieved terminals, and the
  linearized triple patterns, together with attached data properties and
  qualifier patterns, are returned to the agent as a TOON payload.}
  \label{fig:pipeline}
\end{figure}
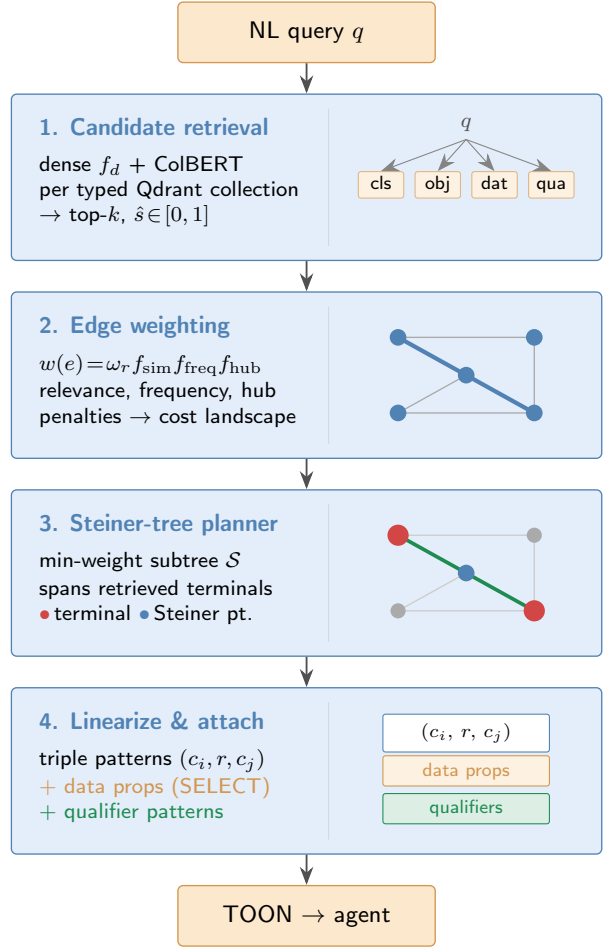

\subsection{Overview}

Query planning proceeds in three stages. A natural-language query is first mapped to ranked candidate schema entities via dual-encoder retrieval over the typed Qdrant\footnote{\url{https://qdrant.tech}} index, with retrieval scores assigned as edge weights on a directed schema graph. A Steiner tree \cite{SteinerProblems} on the undirected projection then yields the minimum-weight join skeleton connecting all retrieved terminals. Data properties and qualifier constraints are attached to the Steiner nodes, producing a structured query-planning package for translation into executable SPARQL by an LLM agent. The workflow is detailed in the following sections and depicted in Figure \ref{fig:pipeline}. All hyperparameters -- including the number of candidates retrieved per collection (k=6), the relation-type base weights, and the property stop list -- were fixed prior to and independently of the evaluation.

\subsection{Schema Graph}
\label{sec:schema-graph}

Although Wikibase organizes its content around items, statements, and qualifiers rather than RDF triples, the endpoint exposes the data through a Wikibase-specific RDF projection. We adopt this RDF view throughout: Schema elements are treated as classes and properties, and the relations between them as directed triples. This allows reasoning about the schema with standard graph machinery while remaining faithful to how queries are actually evaluated against the endpoint.\\

\noindent Let $\mathcal{C}$, $\mathcal{D}$, and $\mathcal{O}$ denote the sets of ontology classes, data properties, and object properties extracted from the MathModDB Wikibase instance. We represent the schema as a directed, weighted graph
\begin{equation}
    \mathcal{G} = (\mathcal{V},\, \mathcal{E}), \qquad \mathcal{V} = \mathcal{C} \cup \mathcal{D},
    \label{eq:schema-graph}
\end{equation}
in which nodes are classes and data properties, and each directed edge $e = (c_i, r, c_j) \in \mathcal{E}$ with $c_i, c_j \in \mathcal{C}$ is labelled by an object property $r \in \mathcal{O}$ and carries a base weight $\omega_r > 0$. The base weight encodes how directly the edge expresses a triple pattern in the underlying RDF projection: we distinguish three relation types with increasing weights, namely (i)~direct object-property links, (ii)~qualifier-mediated object-property links, and (iii)~qualifier-to-data-property usage paths. Higher weights make less direct relations more costly to traverse. To avoid degenerate shortcut paths through generic, high-frequency properties (e.g.,\ \emph{instance of}), we exclude a curated stop list of such properties from $\mathcal{E}$ before any further computation.

\subsection{Candidate Retrieval and Schema Graph Conditioning}
\label{sec:retrieval}

Given a natural-language query $q$, the server first identifies which schema entities are most likely relevant. We use a two-stage retriever that operates separately on each typed Qdrant collection, which are classes, object properties, data properties or qualifier properties.

\paragraph{\textbf{Stage 1: Dense Retrieval}} A dense encoder $f_d$ maps the query and each candidate entity~$e$ into a single vector. Candidates are scored by cosine similarity,
\begin{equation}
    s_d(q, e) = \frac{f_d(q)^\top f_d(e)}{\|f_d(q)\|\,\|f_d(e)\|},
    \label{eq:cosine}
\end{equation}
and the top-$k$ candidates per collection are kept. This stage is tuned for recall and broadly collects candidates.

\paragraph{\textbf{Stage 2: Late-interaction Re-ranking}} The candidates are re-ranked with a ColBERT scorer \cite{ColBert} that compares the query and entity at the token level rather than as single vectors. Let $\mathbf{Q} = [\mathbf{q}_1, \ldots, \mathbf{q}_m]$ and $\mathbf{E} = [\mathbf{e}_1, \ldots, \mathbf{e}_n]$ be the token embeddings of the query and the entity text. The MaxSim score
\begin{equation}
    s_\mathrm{ci}(q, e) = \sum_{i=1}^{m} \max_{j=1}^{n}\, \mathbf{q}_i^\top \mathbf{e}_j
    \label{eq:maxsim}
\end{equation}
lets each query token pick its best-matching entity token before the scores are summed, which improves precision on short, terminology-heavy queries that are typical of schema lookup. To make scores comparable across the four collections, we normalize them within each type to obtain $\hat{s} \in [0, 1]$.

\paragraph{\textbf{Conditioning the schema graph}} The normalized scores are then injected into the schema graph as a query-dependent cost. For each directed edge $e_{ij} = (c_i, r, c_j) \in \mathcal{E}$, we replace the base weight $\omega_r$ by
\begin{equation}
    w(e_{ij}) = \omega_r \cdot f_\mathrm{sim}(e_{ij}) \cdot f_\mathrm{freq}(r) \cdot f_\mathrm{hub}(c_i, c_j),
    \label{eq:edge-weight}
\end{equation}
where the three multiplicative factors penalize, respectively, semantically irrelevant edges, over-represented properties, and edges that pass through high-degree hub nodes. Each factor has a clear intuition. The cost $f_\mathrm{sim}$ penalizes semantically irrelevant edges by combining the retrieval scores of the subject class, the predicate, and the object class into a single edge-level relevance and turning it into a cost,
\begin{equation}
    f_\mathrm{sim}(e_{ij}) = \epsilon + \bigl(1 - \hat{s}_\mathrm{edge}(e_{ij})\bigr)^{\gamma}, \quad \hat{s}_\mathrm{edge} = \alpha_s\,\hat{s}(c_i) + \alpha_p\,\hat{s}(r) + \alpha_o\,\hat{s}(c_j),
    \label{eq:sim-cost}
\end{equation}
so that edges close to the query become cheap and edges far from it become expensive. The constant $\epsilon > 0$ keeps weights strictly positive, $\gamma > 1$ sharpens the contrast between relevant and irrelevant edges, and the weights satisfy $\alpha_s + \alpha_p + \alpha_o = 1$ with $\alpha_p$ dominant, since the predicate is the most informative element of a triple pattern. Next, the factor $f_\mathrm{freq}$ discourages over-represented properties, since properties reused across many triples (e.g.\ generic linking properties) tend to produce ambiguous joins. We penalize them linearly through
\begin{equation}
    f_\mathrm{freq}(r) = 1 + \alpha_f \cdot \frac{\mathrm{freq}(r)}{\max_{r'}\,\mathrm{freq}(r')},
    \label{eq:freq-penalty}
\end{equation}
where \(r'\) is the most frequent property, so that the frequent properties pay larger penalties. Finally, the factor 

\begin{equation}
    f_\mathrm{hub}(c_i, c_j) = 1 + \alpha_h \cdot \delta,\quad \delta \in [0,1]
\end{equation}

is the normalized maximum degree of $c_i$ and $c_j$, discourages paths that funnel through high-connectivity nodes and would otherwise act as cheap but uninformative shortcuts. 

Together, these three factors turn the static schema graph into a query-conditioned cost landscape, without requiring any further model inference at query time.

\subsection{Steiner-Based Join Planning and Qualifier Conditioning}
\label{sec:steiner}

The retrieved classes and data properties form the terminals that the join plan must connect:
\begin{equation}
    T = T_{\mathcal{C}} \cup T_{\mathcal{D}} \subseteq \mathcal{V}, \qquad T_{\mathcal{C}} \subseteq \mathcal{C},\; T_{\mathcal{D}} \subseteq \mathcal{D}.
    \label{eq:terminals}
\end{equation}
On the undirected projection $\tilde{\mathcal{G}}$ of the conditioned schema graph, we look for the minimum-weight connected subtree $\mathcal{S} \subseteq \tilde{\mathcal{G}}$ that spans all terminals in~$T$. This is the classical Steiner tree problem in graphs \cite{SteinerProblems} and produces subgraph $\mathcal{S}$. Intuitively, the Steiner tree is the cheapest backbone in the schema that touches every retrieved entity, and additional non-terminal classes (Steiner points) are added only when they shorten the overall structure. Because the optimization runs on the undirected projection, we restore the original triple direction in a post-processing step by mapping each undirected Steiner edge to its best-matching directed counterpart in $\mathcal{G}$, using property identity and edge weight as tiebreakers.

The resulting subgraph $\mathcal{S}$ is then linearized into candidate triple patterns $(c_i, r, c_j)$, each annotated with the empirical occurrence count taken from the schema index. $\mathcal{S}$ is not a query but a join blueprint and it constrains subsequent SPARQL formulation to schema-valid paths, which reduces the combinatorial space in which an LLM agent must search to produce a correct \texttt{WHERE} clause.

Two further pieces of information are then attached to this blueprint. First, for each Steiner class node $c \in \mathcal{S}$, the data properties that the schema index records as valid for $c$ become projection candidates for the SPARQL \texttt{SELECT} clause. Second, qualifier properties are matched to Steiner subject classes and main properties via the precomputed usage paths, yielding conditional statement patterns for qualified triples. The final output is a structured package containing the Steiner-derived join candidates, the class-conditioned projection fields, and the qualifier-conditioned statement patterns, serialized in TOON format\footnote{https://github.com/toon-format/toon} and handed to the LLM agent. All retrieved classes and data properties are passed as Steiner terminals to the join planner without applying a similarity threshold; precision is delegated to the Steiner optimization, which selects the minimum-weight subgraph over the full terminal set.

\section{Results}

For the evaluation we first analyse the schema-retrieval quality on a set of typical researcher queries, then study two datasets from MathModDB representing different mathematical models from continuum mechanics and enzyme kinetics. The queries for the analysis of the schema-retrieval quality were chosen to cover the main topical areas of the MathModDB schema, targeting different parts of the ontology. The two case studies were selected to represent complementary interoperability scenarios: Case 1 exercises cross-repository linking between a data repository and MathModDB, while Case 2 demonstrates query simplification and federated retrieval within MathModDB and across external knowledge bases including Wikidata.
Together, the quantitative analysis and the case studies demonstrate that coupling an LLM with a curated knowledge base yields both epistemically reliable responses and a level of accessibility that neither component achieves alone. All queries, prompts, LLM chats and other material supporting the results are available in the Supplemental Material section (c.f.~Sec.~\ref{sec:SupplementaryMaterial}).

\paragraph{\textbf{Schema retrieval quality}}
\begin{table}[h]
\centering
\caption{Schema-context provision evaluation: recall at 10 (R@10), precision at 10 (P@10), F1 and mean input-token usage per query, aggregated by topical group.}
\label{tab:scaffold-eval}
\renewcommand{\arraystretch}{1.1}
\setlength{\tabcolsep}{18pt}
\begin{tabular}{l r r r r}
\toprule
\textbf{Topical group} & \textbf{R@10} & \textbf{P@10} & \textbf{F1} & \textbf{\#Tokens} \\
\midrule
Formulations \& equations    & 0.79 & 0.28 & 0.42 & 41\,851 \\
Model transformations        & 0.91 & 0.43 & 0.58 & 27\,229 \\
Tasks \& problem types       & 0.83 & 0.44 & 0.58 & 47\,388 \\
Domain \& provenance         & 0.95 & 0.28 & 0.42 & 32\,218 \\
\midrule
\textbf{Mean}  & \textbf{0.88} & \textbf{0.39} & \textbf{0.53} & \textbf{37\,172} \\
\bottomrule
\end{tabular}
\end{table} 
To assess how reliably \texttt{Explore\_Ontology} exposes the relevant schema scaffold for typical researcher questions, we evaluated the system on 11 natural-language queries spanning four broad areas of the MathModDB schema: model formulations and equations, model transformations, computational tasks and problem types, and domain or provenance information. 

For each query we measure recall at 10 (R@10), precision at 10 (P@10), F1 and token usage per topical group and display the results in Table~\ref{tab:scaffold-eval}. R@10 is defined as the fraction of hand-curated expected schema elements appearing in the agent's output and P@10 as the fraction of returned elements that are in the reference set. The system retrieves k=6 candidates per collection per \texttt{Explore\_Ontology} call by default, however R@10 is reported to provide a more informative assessment of retrieval quality, as k=6 would yield a sample size too small to meaningfully evaluate the vector index. Our approach achieves a mean R@10 of 0.88 across all topical groups. Recall is strongest for domain and provenance queries (0.95) and weakest for formulations and equations (0.79). The mean precision of 0.39 reflects a design choice: since all retrieved schema elements are passed as Steiner terminals to the join planner, the retrieval stage is optimized for recall rather than precision. False-positive retrievals increase the terminal set but do not invalidate the join blueprint, as the Steiner optimization acts as a structural filter downstream, making over-prediction an acceptable trade-off for high recall. The associated token cost is a known consequence of this design; as this work focuses on demonstrating the feasibility of the approach, prompt and retrieval optimization remain future work. Results are stable across repeated runs, with mean P, R, and F1 varying by less than 0.03 across six independent runs (P: 0.39-0.40, R: 0.85-0.88, F1: 0.52-0.53).

\paragraph{\textbf{Case 1: Bridging Dataverse and MathModDB}}
\label{UseCase:1-Dataverse-MMDB}
To evaluate the interoperability between data repositories and MathModDB through the combination of multiple MCP servers, we consider an experimental dataset\footnote{\url{https://doi.org/10.18419/DARUS-5539}} deposited in a Dataverse data repository, which provides concentration trajectories from the enzymatic synthesis of the $\beta$-lactam antibiotic cephalexin, catalyzed by an $\alpha$-amino ester hydrolase (AEH from \textit{X.~campestris}) acting on the substrates PGME and 7-ADCA \cite{Lagerman2021KineticAntibiotics}.

The agent first queried the Dataverse installation via its MCP interface to retrieve the dataset's metadata and files, identifying the four reactant species (PGME, 7-ADCA, CEX, PG) and the catalyzing enzyme. It then accessed MathModDB through its MCP interface, using the ontology-exploration endpoint to obtain the relevant schema scaffolds, specifically the classes for dynamic models, ODEs, and chemical constants, along with their associated direct and qualifier properties. Guided by these scaffolds, the agent issued targeted SPARQL queries against the knowledge graph, retrieving instance-level entries that correspond to the dataset's kinetic structure and resolving their formal definitions and source references.

The combined output from both servers enabled identification of the bi-bi ping-pong mechanism and adjacent variants, which matches the underlying system kinetics. This process demonstrates end-to-end interoperability between the research-data layer and the mathematical-model knowledge layer.

\paragraph{\textbf{Case 2: From complex queries to simple prompts}}
\label{UseCase2:ComplexQueries}
The Stokes Darcy model in MathModDB is a coupled mathematical model describing fluid flow across two distinct physical domains~\cite{Schmalfuss2021}. Typical applications include groundwater–surface water interaction, biomedical flow problems, and industrial filtration.

Retrieving information about the composite Stokes Darcy model from MathModDB  via SPARQL requires detailed knowledge of its schema: the relations linking models to their equations, the properties that identify equations as coupling conditions, and the specific identifiers used throughout the knowledge base. A sample SPARQL query for the coupling conditions of the Stokes Darcy model is available in the Supplementary Materials section. 
When biographical or historical information is requested -- such as the creators of a model -- federated queries across knowledge bases like FactGrid or Wikidata become necessary, since this information is not stored in MathModDB. Queries over multiple knowledge bases demand familiarity with SPARQL federation syntax, endpoint URLs, and the key structures of external databases. 
Our approach of coupling an LLM with a curated knowledge base reduces the user-facing complexity to a natural language question while preserving epistemic rigor. For the coupling condition case, the prompt \texttt{What are the coupling conditions for the Stokes Darcy model in MathModDB?} suffices. For the biographical case, \texttt{Based on MathModDB, who are the creators of the Stokes Darcy model, and where were they born and where did they die?} yields the same result as the federated query.  Internally, the agent first accesses MathModDB via the MCP interface and uses the ontology-exploration endpoint to identify the relevant schema scaffolds, namely the classes for mathematical models, equations, and coupling conditions together with their associated properties. Using these scaffolds as a guide, the agent issues targeted SPARQL queries against the knowledge graph, and for biographical attributes follows external links to Wikidata already present in MathModDB to retrieve additional information without explicit federation.

\section{Discussion}
Our work presents an MCP server for MathModDB that combines a vector-indexed schema retriever, a Steiner-tree-based join planner, and guarded SPARQL execution. The schema-first interface provides LLM agents with ontology-grounded query paths. 

Steiner-tree join planning has been explored in the broader KGQA literature, but our approach integrates it with Wikibase-specific schema extraction -- including qualifier properties and typed entity sets -- dual-encoder schema retrieval with ColBERT re-ranking, and exposure through a stateless MCP interface to LLM agents. Existing SPARQL-MCP approaches such as Dobriy et al.~\cite{Dobriy2026} operate on generic SPARQL endpoints with VoID-based schema exploration and do not address the Wikibase-specific data model or qualifier-mediated triple patterns. The Wikidata MCP server provides broad general-purpose access to a collaboratively edited open knowledge base but lacks a query-conditioned schema planner. Our system adds precisely this layer, targeting a domain-specific, expert-curated resource where schema-valid query paths are essential for scientific reproducibility. 

As shown in \emph{Case 1}, the MCP interface enables cross-repository interoperability, linking research data from Dataverse to mathematical models in MathModDB via natural language. As demonstrated in \emph{Case 2}, the same principle reduces the effort required to access mathematical knowledge within MathModDB. The LLM does not replace the structured knowledge base but uses its schema and data as context for generating targeted SPARQL queries. This allows users to access information through natural-language prompts without having to formulate the underlying queries themselves.

\section{Conclusion and Future Work}
The presented system combines the MathModDB knowledge graph with an LLM through a schema-first MCP interface. The quantitative evaluation shows schema-context retrieval with a mean R@10 of 0.88 across four topical groups. The two case studies demonstrate end-to-end interoperability between heterogeneous data sources -- linking an enzyme-kinetics dataset from Dataverse to its mathematical model in MathModDB (\emph{Case 1}), and simplifying structurally complex federated queries over the Stokes-Darcy model to plain natural-language prompts (\emph{Case 2}). By grounding LLM outputs in epistemically curated knowledge, the system facilitates steps toward XAIR-compliant research outputs: the association of datasets with formal model descriptions, governing equations, and coupling conditions.

The architecture generalizes to any Wikibase-based system exposing a SPARQL endpoint and an indexable schema, making it a reusable integration pattern for the broader Linked Open Data ecosystem, including Wikidata itself. What our solution currently lacks is the reverse process: the extraction of mathematical knowledge from scientific articles and the ingest of extracted model data into MathModDB, which we have identified as the natural next step. Current limitations include the small evaluation set of 11 queries, the absence of a baseline and ablation study, and the lack of a systematic end-to-end reliability analysis. Prompt and retrieval optimization and a controlled usability comparison with the MaRDI portal remain future work.

\section*{Acknowledgments}

This work was funded by the MaRDI project \cite{MaRDI2022} under Deutsche Forschungsgemeinschaft (DFG, German Research Foundation) grant number 460135501, and funded by DFG under Germany's Excellence Strategy -- EXC 2075 -- 390740016. We acknowledge the support by the Stuttgart Center for Simulation Science (SimTech). We also thank the reviewers for their thorough and constructive feedback, which has substantially improved the work.

\newpage

\section*{Supplemental Material Statement}
\label{sec:SupplementaryMaterial}

\subsection*{Source Code}

\begin{itemize}
    \item MathModDB MCP Server
    \begin{itemize}
        \item github repository: \url{https://github.com/MaRDI4NFDI/MathModDB-MCP}
        \item Version release used in this paper: \url{https://doi.org/10.5281/zenodo.22207787}
    \end{itemize}
    \item Schema-retrieval evaluation pipeline: 
    \url{https://github.com/JR-1991/MathModDB-ISWC-Evaluation}
\end{itemize}

\subsection*{LLM Chats/Prompts}\label{sup:chats}

\begin{itemize}
    \item Prompt: \texttt{What are the coupling conditions for the Stokes Darcy model in MathModDB?}  
    \begin{itemize}
        \item Claude Chat share: \url{https://claude.ai/share/71419ffd-fed2-4b51-b36e-b3851cd588bd}
        \item Stable deposited version: \url{https://doi.org/10.5281/zenodo.22208236}
    \end{itemize}\url{https://claude.ai/share/71419ffd-fed2-4b51-b36e-b3851cd588bd}
    \item Prompt: \texttt{Based on MathModDB, who are the creators of the Stokes Darcy model, and where were they born and where did they die?} 
    \begin{itemize}
        \item Claude Chat share: \url{https://claude.ai/share/dcdc556e-c97b-41dd-a58f-5d44c0643fd9}
        \item Stable deposited version: \url{https://doi.org/10.5281/zenodo.22129551}
    \end{itemize}

    \item Prompt: \texttt{Please inspect the dataset 10.18419/DARUS-5539 and gather information on the modelled reaction. Then find a fitting mathematical model for the present reaction system.} 
    \begin{itemize}
        \item Claude chat share: \url{https://claude.ai/share/fe79b9d6-12b4-45d0-a55d-4e56ec1ea704}
        \item Stable deposited version: \url{https://doi.org/10.5281/zenodo.22670774}
    \end{itemize}
    
\end{itemize}

\subsection*{SPARQL Queries}

\begin{itemize}
\item SPARQL query to determine the coupling conditions of the Stokes Darcy model:\\ \url{https://gist.github.com/bjschembera/b555e3c1af5aece36c297e01423725dd}
\item Federated SPARQL query to find biographical information about the creators of the Stokes Darcy model: \url{https://gist.github.com/bjschembera/1700f34ed6e59d89e519f06ab5384c5b}
\end{itemize}

\section*{Declaration of use of Generative AI}

In the preparation of this work, the following generative AI tools were used:

\begin{itemize}

    \item \textbf{Claude Opus 4.7 \& Sonnet 4.6} has been used for language polishing, stylistic refinement, and feedback on the structure and organization of paragraphs. It has also been used for the creating  figures~\ref{fig:stack} and~\ref{fig:pipeline}.
    \item \textbf{Grammarly} has been used for proofreading, grammar and spelling checks, and to obtain feedback on reader reactions and clarity.

\end{itemize}

\noindent All AI-assisted outputs were reviewed and edited by the authors, who take full responsibility for the content of the work.

\appendix

\bibliography{references}

\end{document}